\documentclass[sigconf]{acmart}

\makeatletter

\newcommand\addauthornote[1]{%
  \if@ACM@anonymous\else
    \g@addto@macro\addresses{\@addauthornotemark{#1}}%
  \fi}

\newcommand\@addauthornotemark[1]{\let\@tmpcnta\c@footnote
   \setcounter{footnote}{#1}\addtocounter{footnote}{-1}
    \g@addto@macro\@currentauthors{\footnotemark\relax\let\c@footnote\@tmpcnta}}

\makeatother
\usepackage{booktabs} 
\usepackage{makecell}
\usepackage{amsmath}
\usepackage[nopar]{lipsum}
\usepackage{hyperref}
\usepackage{multirow}
\usepackage{booktabs,makecell,tabularx}

\usepackage{siunitx}
\usepackage{adjustbox}
\usepackage{array,booktabs}
\usepackage{graphicx}
\usepackage{epstopdf}

\usepackage{array}

\newcolumntype{L}[1]{>{\raggedright\let\newline\\\arraybackslash\hspace{0pt}}m{#1}}
\newcolumntype{C}[1]{>{\centering\let\newline\\\arraybackslash\hspace{0pt}}m{#1}}
\newcolumntype{R}[1]{>{\raggedleft\let\newline\\\arraybackslash\hspace{0pt}}m{#1}}
\makeatletter
\DeclareRobustCommand\onedot{\futurelet\@let@token\@onedot}
\def\@onedot{\ifx\@let@token.\else.\null\fi\xspace}

\newcommand{\etal}{\textit{et al}.}
\begin{document}
\title{iSPA-Net : Iterative Semantic Pose Alignment Network}


\author{Jogendra Nath Kundu}
\authornote{Equal contribution}
\affiliation{
  \institution{Indian Institute of Science}
  \city{Bengaluru}
  \country{India}}
\email{jogendrak@iisc.ac.in}

\author{Aditya Ganeshan}
\affiliation{
  \institution{Indian Institute of Science}
  \city{Bengaluru}
  \country{India}}
\email{adityaganeshan@gmail.com}
\addauthornote{1}

\author{Rahul M V}
\affiliation{
  \institution{Indian Institute of Science}
  \city{Bengaluru}
  \country{India}}
\email{rahulmv.cs14@rvce.edu.in}
\addauthornote{1}

\author{Aditya Prakash}
\affiliation{
  \institution{Indian Institute of Science}
  \city{Bengaluru}
  \country{India}}
\email{adityaprakash229997@acm.org}

\author{R. Venkatesh Babu}  
\affiliation{
  \institution{Indian Institute of Science}
  \city{Bengaluru}
  \country{India}}
\email{venky@iisc.ac.in}



\begin{abstract}
Understanding and extracting 3D information of objects from monocular 2D images is a fundamental problem in computer vision. In the task of 3D object pose estimation, recent data driven deep neural network based approaches suffer from scarcity of real images with 3D keypoint and pose annotations. Drawing inspiration from human cognition, where the annotators use a 3D CAD model as structural reference to acquire ground-truth viewpoints for real images; we propose an iterative Semantic Pose Alignment Network, called \textit{iSPA-Net}. Our approach focuses on exploiting semantic 3D structural regularity to solve the task of fine-grained pose estimation by predicting viewpoint difference between a given pair of images. Such image comparison based approach also alleviates the problem of data scarcity and hence enhances scalability of the proposed approach for novel object categories with minimal annotation. The fine-grained object pose estimator is also aided by correspondence of learned spatial descriptor of the input image pair. The proposed pose alignment framework enjoys the faculty to refine its initial pose estimation in consecutive iterations by utilizing an online rendering setup along with effectiveness of a non-uniform bin classification of pose-difference. This enables \textit{iSPA-Net} to achieve \textit{state-of-the-art} performance on various real image viewpoint estimation datasets. Further, we demonstrate effectiveness of the approach for multiple applications. First, we show results for active object viewpoint localization to capture images from similar pose considering only a single image as pose reference. Second, we demonstrate the ability of the learned semantic correspondence to perform unsupervised part-segmentation transfer using only a single part-annotated 3D template model per object class. To encourage reproducible research, we have released the codes for our proposed algorithm\footnote{Link to repository: \href{https://github.com/val-iisc/iSPA-Net}{https://github.com/val-iisc/iSPA-Net}}.
 
\vspace{34mm}
\end{abstract}

%
%


\copyrightyear{2018} 
\acmYear{2018} 
\setcopyright{othergov}
\acmConference[MM '18]{2018 ACM Multimedia Conference}{October 22--26, 2018}{Seoul, Republic of Korea}
\acmBooktitle{2018 ACM Multimedia Conference (MM '18), October 22--26, 2018, Seoul, Republic of Korea}
\acmPrice{15.00}
\acmDOI{10.1145/3240508.3240650}
\acmISBN{978-1-4503-5665-7/18/10}

\maketitle
\section{Introduction}

Although human creativity has led us to create ingenious, highly varied designs for objects like cars, furnitures etc., the intrinsic structure of different instances of an object category are similar. By having a clear understanding of the 3D structural regularities despite the diverse intra-class variations, machines can gain a holistic understanding of their environment. A task that goes hand-in-hand with this aim is object viewpoint estimation, the task of estimating the 3D pose of an object from a 2D image. In this work, we explore whether the relation between 3D structural regularities and viewpoint can be exploited for fine-grained viewpoint estimation.
In our daily routine we see objects from various viewpoints overtime. This helps human perception to manipulate one view of the object with another view of the same object class already perceived in temporal proximity \cite{perry2006spatial}. In fact, to accurately annotate object viewpoint in various datasets \cite{xiang2016objectnet3d,xiang2014beyond}, the annotators have the provision to compare the object in image, to a reference 3D model projection at different viewpoints. Inspired from both the human cognition and recent advances in Deep Learning, we take a novel direction to solve the task of object pose estimation using a single 3D template model as structural reference. A fine-grained iterative Semantic Pose Alignment Network, named \textit{iSPA-Net}, is developed to efficiently model a view-planning strategy, which can iteratively match pose of a given 3D template to the pose of object in an RGB image.

\begin{figure*}[!t]
\centering    
	\includegraphics[width=0.95\linewidth]{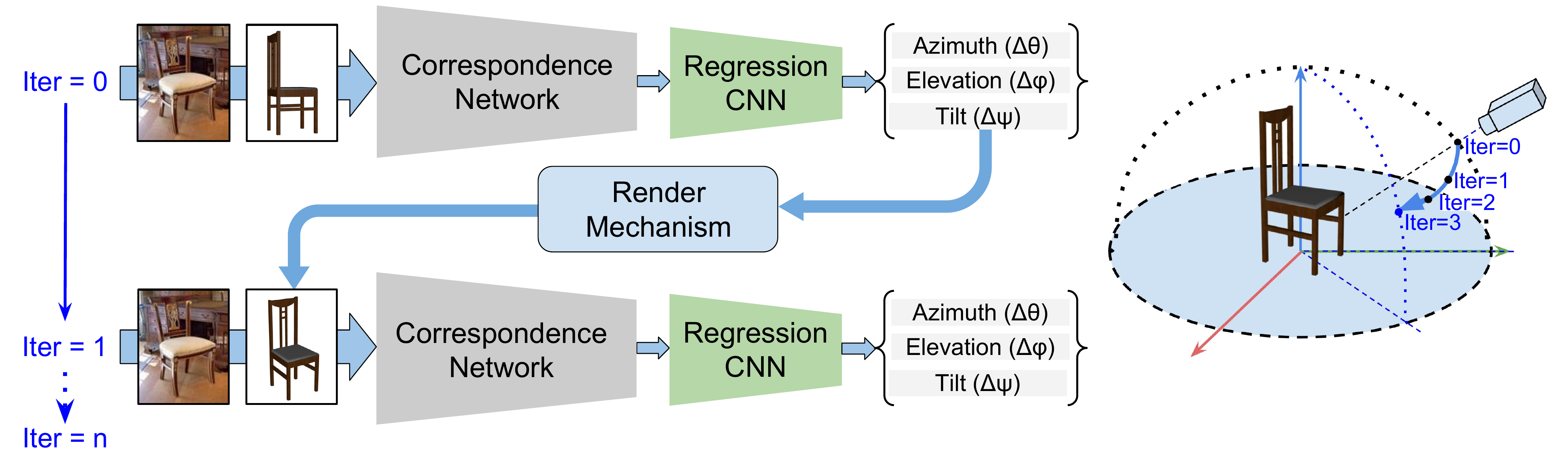}
	\caption{Illustration of the proposed pipeline. At each iteration, \textit{iSPA-Net} takes a pair of real and graphically rendered image to predict the difference in viewpoint. In order to improve pose alignment, a new synthetic image is rendered from the previously estimated viewpoint in each iteration. The figure on right shows the trajectory of viewpoints for the iterative alignment.} 
	\label{fig:fig_1}    
\end{figure*}

Our proposed approach performs object viewpoint estimation by semantic alignment of pose of a reference 3D model projection to the pose of object in a given natural image. To realize this, we employ a novel CNN architecture which takes the pair of images as input to estimate a difference in object viewpoint. The proposed architecture consists of two major parts, a correspondence network followed by a pose-estimator network. The correspondence network computes a correspondence tensor capturing information regarding spatial displacement of object parts between the two input images. Following this, the pose-estimator network infers the difference in object viewpoint. To obtain an absolute viewpoint prediction of the given real object, we add the predicted viewpoint difference to the known viewpoint of the synthetically rendered image. To further achieve improvements on pose estimation result, we introduce an iterative estimation pipeline where, at each iteration the reference 3D template model is rendered from the viewpoint estimate of the previous iteration (see Figure \ref{fig:fig_1}).

Object viewpoint estimation has been attempted previously using Deep Convolutional Networks (CNNs) in \cite{tulsiani2015viewpoints,su2015render,xiang2014beyond}, but these approaches have several drawbacks. Works such as RenderForCNN \cite{su2015render} etc., solve this task by employing a CNN to directly regress the pose parameters. Due to lack of structural supervision, such methods do not learn the underlying structural information in an explicit manner. Other works such as 3D-INN  \cite{wu2016single} and DISCO~\cite{li2017deep} propose to primarily predict an abstract 2D skeleton map for a given RGB image. While performance of such works for keypoint estimation is remarkable, they deliver sub-optimal results for viewpoint estimation. This is understandable as the central focus of such works is keypoint estimation, which does not encourage \textit{fine-grained} viewpoint estimation explicitly.

On the other hand, \textit{iSPA-Net} enjoys advantages of a number of key design choices as compared to the previous approaches. The iterative alignment pipeline is one of the key novel features of the proposed pose estimation framework. This enables \textit{iSPA-Net} to refine its pose estimation results in successive iterations using the online rendering pipeline in contrast to previous state-of-the-arts. Another key-feature of \textit{iSPA-Net} is that instead of inferring absolute viewpoint, we estimate viewpoint difference between a pair of image projections. This enables us to utilize $ ^nC_2$ training examples with only $n$ unique real images having pose annotation. Due to this, we are able to efficiently utilize the available data, which in the case of viewpoint estimation, is very scarce due to high cost of manual annotation. Previous works like RenderForCNN \cite{su2015render} and 3D-INN \cite{wu2016single} tackle the scarcity of data by generating synthetic data in abundance (millions in \cite{su2015render,wu2016single}) along with the available real samples. However, it is know that statistics of synthesized images are different from those of real images \cite{kundu2018adadepth}, which often leads to sub-optimal performance on real data.
To explicitly enforce fine-grained pose estimation, \textit{iSPA-Net} uses a non-uniform binning classifier for estimating the viewpoint difference. Bins vary for coarse to fine partitions as the viewpoint difference varies from large to small. By enforcing the network to progressively improve its precision as viewpoint difference decreases, we ensure that pose estimation is improved in successive iterations (Section \ref{section:36}). 


Several experiments are conducted on real image datasets to demonstrate effectiveness of the proposed approach (Section  \ref{section:4}). We achieve \textit{state-of-the-art} results on two viewpoint estimation datasets, namely PASCAL 3D+ \cite{xiang2014beyond}, and Objectnet3D \cite{xiang2016objectnet3d}.
We also show two diverse applications of \textit{iSPA-Net}: Firstly, we present the task of active object viewpoint localization (Section  \ref{section:41}), where, given a reference image, the agent must relocate its position so that pose of object in the camera feed matches with the pose of object in a given reference image. This has a wide range of applications in industrial and warehousing setups. For instance, automated cataloging of all the chairs in a showroom from a particular viewpoint referred in a reference image, using a camera mounted drone. Secondly, we utilize the learned semantic correspondence of \textit{iSPA-Net} for performing unsupervised transfer of part segmentation on various objects using a single part-annotated 3D template model (Section  \ref{section:42}).


To summarize, our main contributions in this work include: An approach for object viewpoint estimation, which (1) by its iterative nature enables accurate fine-grained pose estimation, and (2) by predicting difference of viewpoint between objects in a image pair, alleviates the data bottleneck. We also show (3) \textit{State-of-the-art} performance in object viewpoint estimation on various datasets. Finally,(4) multiple applications of the proposed approach are shown such as Active Object Viewpoint Localization on synthetic data and Unsupervised Part-Segmentation on real data.

\section{Related work}

\textbf{3D structure inference from 2D image:} One of the fundamental goals of computer vision is to infer 3D structural information of objects from 2D images. A few recent works attempt to solve this in an unsupervised way using projection transformation by employing special layers with deep convolutional network architecture \cite{rezende2016unsupervised,yan2016perspective,choy20163d,li2015joint}.  Kalogerakis \etal \cite{liu2016sift} follows a similar approach to transfer object part annotations from multiple 2D view projections to 3D mesh models. Before deep era, projected location of unique object parts or semantic key-points were also explored to infer 3D viewpoint of objects from RGB images \cite{aubry2014seeing,lim2014fpm, su2014estimating,xu2016data}. Earlier methods employed hand-engineered local descriptors like SIFT or HOG \cite{aubry2014seeing,liu2016sift,taniai2016joint,berg2005shape} to represent semantic part structures useful for viewpoint estimation. Recent works, such as \cite{li2017deep,wu2016single}, extract 3D structural information by predicting 3D keypoint locations as an intermediate step while estimating 2D keypoints. However, as projection of 2D keypoints to 3D space, being a ill-posed problem, is prone to be erroneous, the 3D structural information may not be precise. Additionally, due to estimation of only a few keypoints, the extracted 3D structural information might not be suitable for high precision task such as fine-grained viewpoint estimation.

\begin{figure*}[!t]
\centering    
\includegraphics[width=0.88\linewidth]{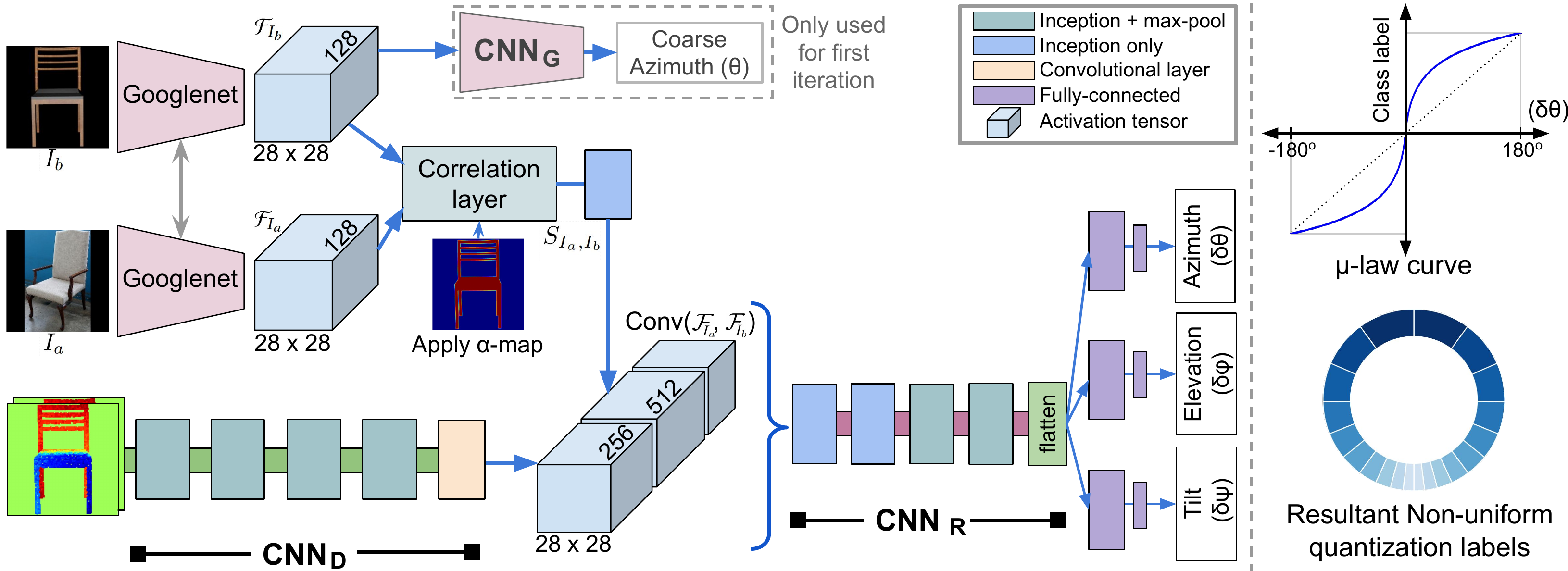}
 	\caption{Left: Illustration of the proposed \textit{iSPA-Net} architecture with connections between the individual CNN modules (Section \ref{section:3}). Right: $\mu$-law curve along with the resultant non-uniform bin-quantization used for prediction of angle-difference.
 	} 
 	\label{fig:fig_2}    
\end{figure*}

Multitudes of work, such as \cite{schmidt2017self,han2017scnet,yu2018hierarchical,choy2016universal}, propose use of CNNs for learning semantic correspondence between images. Universal Correspondence Network \cite{choy2016universal} proposes an optimization technique for deep networks to learn robust spatial correspondence by efficiently designing an active hard-mining strategy and a convolutional spatial transformer.

\vspace{1mm}
\noindent
\textbf{Object viewpoint estimation:} \hspace{1mm} There are many recent works which use deep convolutional networks for object viewpoint estimation \cite{poirson2016fast,mahendran20173d}. RenderForCNN \cite{su2015render} was one of the first to use deep CNNs in an end-to-end approach solely for 3D pose estimation. They synthesized rendered views from 3D CAD models with random occlusion and background information to gather enough labeled data required to train a deep network. However, such approach may not be applicable for novel object categories as it requires a large repository of 3D CAD models for creating labeled training data. Following an entirely different approach, 3D Interpreter Network (3D-INN) \cite{wu2016single} proposed an convolutional network to predict 2D skeleton locations. They recover the 3D joint-point locations and the corresponding 3D viewpoint from the estimated 2D joints by minimizing the reprojection error. However, such skeleton based approach heavily rely on correctness of keypoint annotation data for natural images, which is especially noisy for occluded keypoints. Tulsiani \etal \cite{tulsiani2015viewpoints} explored the idea of coarse to fine level view-point estimation. An end-to-end approach to directly classify discretized viewpoint angles can only be used to perform coarse level 3D view estimation. Such methods extract global structural features, whereas for fine-grained pose estimation, spatial part-based keypoints plays a crucial role. Hence, in our proposed approach, the iterative alignment frameworks starts from a coarse level viewpoint estimation followed by fine-level alignment of a structural template model by enforcing explicit local descriptor correspondence.

\vspace{1mm}
\noindent
\textbf{3D model retrieval for viewpoint estimation:} \hspace{1mm} A cluster of previous arts exist which attempt to estimate 3D structure of object by aligning a retrieved matching CAD model using 2D RGB images \cite{aubry2014seeing,xiang2014beyond,massa2016deep} and also additional depth information \cite{bansal2016marr,gupta2015inferring}. But performance of these methods highly rely on the style of the retrieved CAD model due to the high intra-class variations. Additionally, such works cannot be adapted for the proposed task of active object viewpoint localization as they require CAD model of object for alignment, which may not be available in real life applications.

\section{Approach} \label{section:3}

In this section, we explain details of the proposed pose alignment pipeline (refer Figure \ref{fig:fig_2} for an overview). The architecture is inspired from a classical computer vision setup~\cite{philbin2007object} with the addition of specially designed modules which make the pipeline fully differentiable for end-to-end training. A classical pipeline would typically start with extraction of useful local descriptors (e.g. SIFT, HOG) or spatial representations for both the given images. Then, the part-based spatial features are matched between the two images to acquire a correspondence map, which is then used to infer pose shift or geometrical parameters for alignment (e.g. RANSAC). Our architecture also follows a similar pipeline to align pose of the 3D template model over a given natural image. The different components of our proposed approach are presented in following subsections.

\subsection{View-invariant feature representation, ($\mathcal{F}_{I})$}\label{section:31}

As discussed earlier, we first focus on extraction of useful local features, which can result in efficient local-descriptor correspondence. As shown in Figure \ref{fig:fig_2}, the network takes two input images ($I_a$ and $I_b$) through a Siamese architecture with shared parameters, and outputs corresponding spatial feature maps. Here, $I_a$ is an image containing the object of interest whereas $I_b$ is a rendered RGB image generated from a 3D template model with known viewpoint parameters i.e. azimuth ($\theta$), elevation ($\phi$), and tilt or in-plane rotation ($\psi$). We represent the output feature map of the this network, which is analogues to local descriptors used in classical setup, as $\mathcal{F}_{I_a}$ and $\mathcal{F}_{I_b}$. 
To learn spatial representations essential for part alignment, we use correspondence contrastive loss, as presented in \cite{choy2016universal}. Let  $\mathbf{x_i}$ and $ \mathbf{x'_i} $ represent spatial locations on $I_a$ and $I_b$ respectively. Then, the contrastive loss can be defined as,

\begin{eqnarray}\label{eqn:1}
L &=& 
	\dfrac{1}{2N}\sum_{i}^{N}s_i{\lVert \mathcal{F}_{I_a}(\mathbf{x}) - \mathcal{F}_{I_b}(\mathbf{x'}) \rVert}^2 + \nonumber \\ & &
	(1-s_i) \max{(0, \; m-{\lVert \mathcal{F}_{I_a}(\mathbf{x}) - \mathcal{F}_{I_b}(\mathbf{x'}) \rVert}^2)}
\end{eqnarray}

where $N$ is the total number of pairs, $s_i = 1$ for positive correspondence pairs, and $s_i = 0$ for negative correspondence pairs. In Section \ref{section:35}, we describe how positive correspondence pair are acquired between a pair of images, $(I_a,I_b)$. 

\vspace{2mm}
\noindent
\textbf{Correspondence Network Architecture:} \hspace{1mm}
For the Siamese network, we employ a standard GoogLeNet~\cite{szegedy2015going} architecture with imagenet pretrained weights. To obtain spatially aligned local features $\mathcal{F}_I$, we use a convolutional spatial transformation layer, as proposed in UCN \cite{choy2016universal}, after $pool4$ layer of GoogLeNet architecture. The convolutional spatial transformation layer greatly improves feature correlation performance by explicitly handling scale and rotation parameters for efficient correspondence among the spatial descriptors of $I_a$ and $I_b$.

\subsection{Correspondence map and Disparity network ($CNN_D$)}\label{section:32}

The output feature map obtained, $\mathcal{F}_{I_a}$ and $\mathcal{F}_{I_b}$, are of size $h \times w \times d$ tensors. After L2 normalization, these can also be considered as a $d$-dimensional spatial descriptors for each location $(i, j)$, i.e. $\mathcal{F}_{I_a}(i, j) \in \mathbb{R}^d$. A feature correlation layer is formulated to get spatial correlation map for all location pairs $(i, j)$, $(i',j')$ covering the full resolution $h \times w$ of feature maps $\mathcal{F}_{I_a}$ and $\mathcal{F}_{I_b}$ respectively. The pairwise feature correlation for any location pair $(i, j)$ and $(i',j')$ is computed as:
$$
S_{I_a,I_b}(\mathcal{F}_{I_a}(i,j), \; \mathcal{F}_{I_b}(i',j')) = \dfrac{ \mathcal{F}_{I_a}(i,j)^T \mathcal{F}_{I_b}(i',j')}
{\sqrt{ \sum_{k, l}(\mathcal{F}_{I_a}(k,l)^T \mathcal{F}_{I_b}(i',j'))^2}}
$$

In the above formulation, dot product between the l$2$ normalized spatial descriptors is taken as a measure of correlation. The correlation maps are also normalized for each location $(i', j')$ of the input feature map $\mathcal{F}_{I_b}$ across all locations $(k,l)$ of the other feature map, $\mathcal{F}_{I_a}$. Due to this normalization, ambiguous correspondences having multiple high correlation matches with the other spatial feature map are penalized. Such normalization step is in line with the traditionally used second nearest neighbor test proposed by Lowe \etal ~\cite{lowe2004distinctive}. The final resultant tensor $S_{I_a,I_b}$ is of size $h \times w \times (h*w)$ representing location wise spatial matching index of part-based descriptors for a given pair of input images $(I_a, I_b)$. Here, both correlation and normalization steps are clearly differentiable with simple vector and matrix operations, thus enabling end-to-end training of the pose alignment framework. 

To gain a compact and fused representation of the correspondence, $S_{I_a,I_b}$ is further passed through a single inception module. Finally, we apply the down-sampled $\alpha$ map (transparency map) obtained from the rendering mechanism of $I_b$ to every feature channel of the correspondence tensor.
For further processing, the correspondence feature, $S_{I_a,I_b}$ is then concatenated with the spatial features $\mathcal{F}_{I_a}$ and $\mathcal{F}_{I_b}$ after processing through some convolutional layers. This combines the part-based local descriptor representation of $\mathcal{F}_{I_a}$ and $\mathcal{F}_{I_b}$  with the corresponding spatial shift ($S_{I_a,I_b}$) obtained between $I_a$ and $I_b$. Next, a disparity-map between a stereo pair of rendered image, $I_b$ with $I_b'$ is computed. Here, $I_b'$ is generated by considering a minimal shift of $10^{\circ}$ in both azimuth and elevation angle at the absolute viewpoint of object in $I_b$. The raw disparity map is fed to a small disparity network $CNN_D$, allowing the network to exploit useful 3D information of the template object. Finally, as shown in Figure \ref{fig:fig_2}, the extracted representation is merged with the previously acquired concatenated tensor to obtain the input tensor to the Pose-estimator network $CNN_R$. 

Note that only appearance-based disparity is used, which can be easily captured without access to the 3D model, as the case might be in real-world scenario. In Section~\ref{ex:ablations}, we show ablation on our proposed architecture and demonstrate the utility of each of the aforementioned components.

\subsection{Pose-estimator network, ($CNN_R$)}\label{section:33}

The merged representation, containing output of $CNN_D$ along with correlation map and the view invariant feature tensor, is then passed as input to the Pose-estimator Network, $CNN_R$. The network is trained to predict the viewpoint difference $(\theta_b - \theta_a, \phi_b - \phi_a, \psi_b - \psi_a)$ = $(\delta\theta, \delta\phi, \delta\psi)$ = $\Delta$ between the viewpoints of object in image $I_a$ and $I_b$. We model the viewpoint difference $\Delta$ as an $n$-bin classification problem. The $n$ bins for classification are formed using $\mu$-Law quantization of $\Delta$, which is explained in section \ref{section:35}. The network is trained using the Geometric Structure Aware Loss Function, proposed in \cite{su2015render}. Further, we introduce an auxiliary task of predicting the absolute azimuth angle $\theta_a$ of real object in image $I_a$, and experimentally evaluate its utility.  

\subsection{Iterative pipeline and viewpoint classifier ($CNN_G$) }\label{section:34}
 
 The proposed method, currently consist of a pipeline which, given two input images $I_a$ and $I_b$, estimates the viewpoint difference between them. To estimate the viewpoint of object in an image $I_a$, it is passed along with another image $I_b$ containing a rendered 3D template object at a known viewpoint parameters $(\theta_b, \phi_b, \psi_b )$. The viewpoint in $I_a$ is estimated to be $(\theta_b + \hat{\delta\theta} , \phi_b+ \hat{\delta\phi}, \psi_b + \hat{\delta\psi})$ using the predicted difference $(\hat{\delta\theta}, \hat{\delta\phi}, \hat{\delta\psi})$. However, for fine-grained pose estimation, we propose the following iterative pipeline. Consider at iteration $i$, the viewpoint of $I_b^i$ is $(\theta_b^i, \phi_b^i, \psi_b^i )$ and the predicted viewpoint difference is $(\hat{\delta\theta^i}, \hat{\delta\phi^i}, \hat{\delta\psi^i})$. Now, as it is possible to render an image $I_b^{i+1}$ with the viewpoint $(\theta_b^i + \hat{\delta\theta^i} , \phi_b^i+ \hat{\delta\phi^i}, \psi_b^i + \hat{\delta\psi^i})$, for iteration $i+1$, the pair $I_a$ and $I_b^{i+1}$ acts as input pair, thereby allowing the network to perform fine-grained viewpoint estimation. This process is continued until the estimated viewpoint difference is below some threshold $\mathcal{T} = (\tau_\theta, \tau_\phi, \tau_\psi)$, or when the iteration limit $n$ is reached. 
 
While it is possible to randomly initiate the viewpoint of 3D template object for the first iteration, we develop a more structured approach by employing a small Viewpoint Classifier Network, $CNN_G$. This network takes $\mathcal{F}_{I_a}$ as input to provide a coarse viewpoint estimate of the object in image $I_a$. This estimate is used as initial viewpoint of $I_b$ in the iterative pipeline. By using this coarse estimate, the number of iterations required to reach the threshold $\mathcal{T}$ (on an average) is reduced significantly. $CNN_G$ is a shallow network comprising of only three convolutional layers with a final classification layer. It is trained separately for a 16-way classification of only azimuth angle ($\theta_a$) using the Geometric Structure Aware Loss proposed in \cite{su2015render}.

\subsection{$\mu$-Law Quantization of angle difference ($\Delta$) }\label{section:36}
Ideally, for fine grained pose alignment, the model should be precise in its prediction  viewpoint difference $\hat{\Delta}$ for all $\Delta$ large and small range. However, such a model would require high capacity, as well as large quantity of data. Instead, we propose an alternate solution, where the model is biased to have precise predictions when $\Delta$ is small, and only approximate prediction when $\Delta$ is large. When used in an iterative setup, such a model would reduce $\Delta$ at each iteration, leading it to have improved precision in successive iterations. Hence, this model can sufficiently address the task of fine grained pose alignment, without facing the capacity and data bottleneck of the \textit{ideal} model.
 
In \textit{iSPA-Net}, we realize such a bias in the system by introducing a non-uniform binning for the output of Pose-estimator network, ($CNN_R$). Instead of a uniform $360$-bin $\delta\theta$ classification, we perform a $n$-bin $\delta\theta$ classifier, where finer bins are allocated to lower $\delta\theta$ range and coarser bins for higher $\delta\theta$ range. We use the $\mu$-Law curve to label each $\delta\theta$ to a bin. The right section of Figure \ref{fig:fig_2}, shows the $\mu$-law curve and a representative labeling of $\delta\theta$ into 20 bins, using the proposed formulation. The $\mu$-law equation used to obtain the non-uniform binning of angle-difference can be written as:
 $$
 Bin(\Delta)  = sign(\Delta) * \ln(1 + \mu \lvert \Delta \rvert) /  \ln(1+\mu); -180 < \Delta < 180.
 $$
 Similarly, we use the inverse function $Bin^{-1}(b) = sign(b)(1/\mu)((1 + \mu)^{ \lvert b \rvert} -1)$ to obtain $\hat{\Delta}$ prediction from the bin-classifier.
 
 Another advantage of this approach is that while it provides performance comparable to \textit{iSPA-Net} with additional recurrent layers ( refereed as \textit{iSPA-Net$_R$}) and normal uniform binning, its training regime is substatially simpler, and faster. Training is simpler due to the single iteration training of \textit{iSPA-Net}, compared to the online iterative training required for \textit{iSPA-Net$_R$}. 
Hence, \textit{iSPA-Net} is truly iterative only during the validation or test setup. 
\subsection{Data preparation}\label{section:35}

We select a single 3D template model $M_c$ for each class of object $c \in C$ from ShapeNet dataset \cite{shapenet2015}. 
Using a modified version of the rendering pipeline presented by  \cite{su2015render}, the selected template model $M_c$ is rendered at various viewpoints to create sample images $I_b(\theta_b, \phi_b, \psi_b )$. 
By pairing these images randomly with real images $I_a(\theta_a, \phi_a, \psi_a )$, we create training samples for our pose alignment framework. Our network takes $(I_a,I_b)$ as input image pair and the corresponding ground truth is obtained as $(\delta\theta, \delta\phi, \delta\psi)$ i.e. $(\theta_b - \theta_a, \phi_b - \phi_a, \psi_b - \psi_a)$. Note that, our reliance on synthetic data is minimal. For efficient offline training, where rendered images $I_b$ are paired with real images $I_a$, we use only 8,000 renders of  a \textit{single} 3D template model. This is in sharp contrast to works such as \cite{su2015render,li2017deep}, which use millions of synthetic images in their training pipeline.

\begin{figure}[!t]
\centering    
\includegraphics[width=1.0\linewidth]{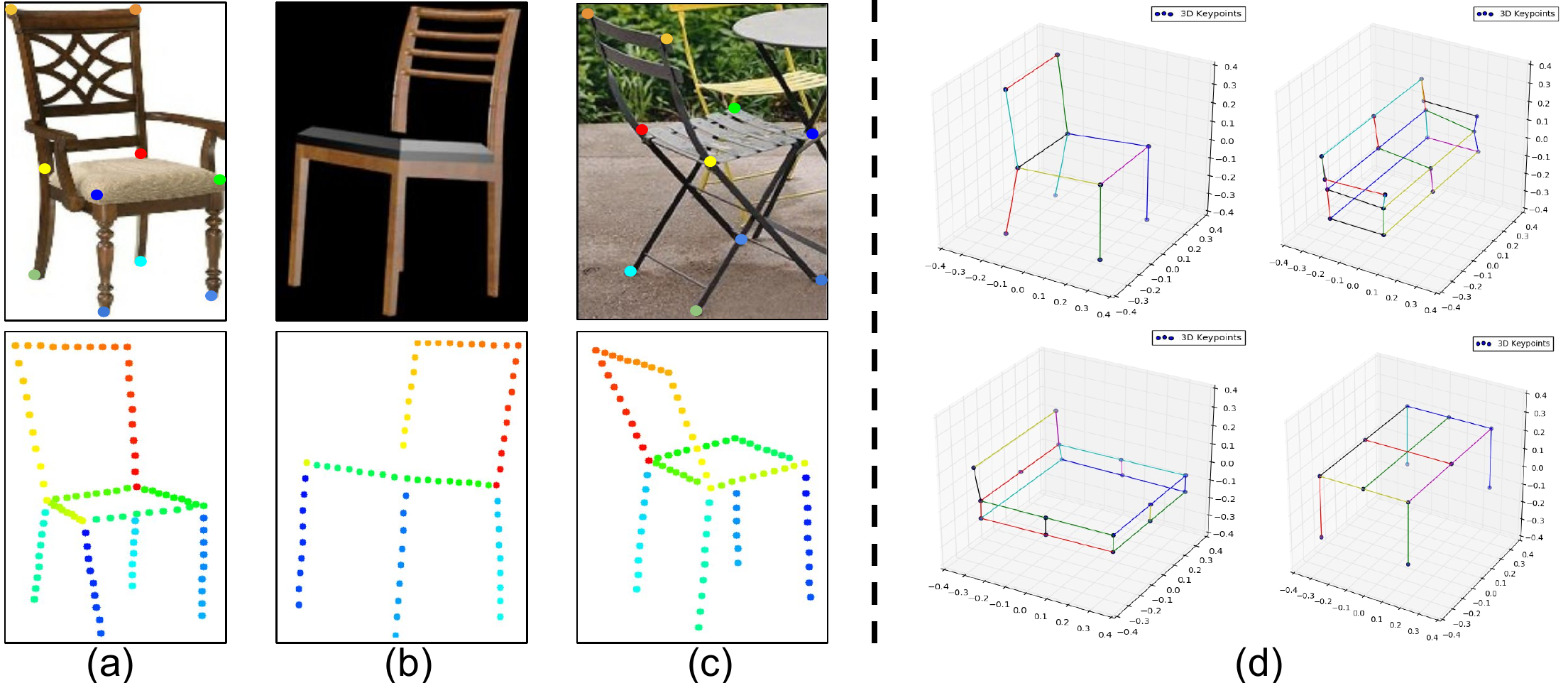}
 	\caption{Left: Top row of (a) and (c) show examples from Keypoint-5 dataset and top row of (b) shows a synthetic rendered sample. Bottom row  in all three depicts the generated 2D skeletal frames. Right (d): Manually annotated 3D skeletal model for the \textit{single} 3D template model of each object category.  
 	} 
 	\label{fig:fig_3}    
\end{figure}

While this information is sufficient to train the network in an end-to-end fashion, to further improve our view-invariant feature representation $\mathcal{F}_I$, we use the loss function given by equation \ref{eqn:1}, which requires dense correspondence annotations between image pair $(I_a,I_b)$. For generating such correspondence information, we use automated processing of annotations provided in the Keypoint-5 dataset, released by Wu~\etal~\cite{wu2016single}. We use the 2D skeletal representation, which is based on annotation of sparse 2D skeletal keypoints of real images. As shown in Figure \ref{fig:fig_3}, the sparse 2D keypoints are annotated on real images at important joint locations such as leg ends, seat joint etc. For each image sample $I_a$, we join these sparse keypoints in a fixed order, to obtain the corresponding 2D skeletal frame $S_a$ (second row of Figure \ref{fig:fig_3}: a and c). 
To generate a similar skeletal frame for our rendered 3D object templates in $I_b$, we manually annotate sparse 3D keypoints for our template objects, as shown in Figure \ref{fig:fig_3}d. Using the projection of these 3D keypoints, based on the viewpoint parameters, we generate 2D keypoints for any rendered image $I_b$ and also the corresponding  2D skeletal frame $S_b$ in a similar fashion. 
Now, by pairing points along this generated skeletal frames $(S_a,S_b)$ of any image pair $(I_a,I_b)$, 2D keypoints on $I_a$ can be matched to the corresponding keypoints on image $I_b$. By creating multiple such pairs, we generate dense correspondence set for any image pair $(I_a,I_b)$ which is then used to improve the performance of local descriptor $\mathcal{F}_I$.

Figure \ref{fig:fig_3} shows some qualitative examples of generated annotations for our training. We employ various methods to prune and reform our generated annotations, such as depth based pruning (for $I_b$), and other methods (details provided in supplementary). Although, sometimes the generated annotations are not accurate (Figure \ref{fig:fig_3}.c), the correspondence model is able to learn improved view invariant local descriptors $\mathcal{F}_I$, due to the presence on ample amount of correct noise-free annotations.


\section{Experiments}\label{section:4}

In this section, we compare the proposed approach \textit{iSPA-net} with other \textit{state-of-the-art} methods for viewpoint estimation. We also examine the performance improvement caused by the different design decisions for \textit{iSPA-Net}. 

\vspace{2mm}
\noindent
\textbf{Datasets and Metrics} \hspace{1mm}
We empirically demonstrate \textit{state-of-the-art} performance when compared to several other methods, on two public datasets, namely Pascal3D+\cite{xiang2014beyond}, and ObjectNet3D\cite{xiang2016objectnet3d}. We evaluate our performance for the task of object viewpoint estimation.

Performance in object viewpoint estimation is measured using \textit{Median Error} ($MedErr$), and \textit{Accuracy at $\theta$} ($Acc_\theta$), which were introduced by Tulsiani \etal~\cite{tulsiani2015viewpoints}. $MedErr$ is measured in terms of degrees. As our approach aims to perform fine-grained object viewpoint estimation, we show $Acc_\theta$ with smaller $\theta$ as well. This stronger metric requires higher precision in estimation of pose and highlights our models utility for fine-grained pose estimation. Finally, we show $Acc_{\theta}$ vs $\theta$ plots to concretely establish our superiority in fine-grained pose estimation.  

\vspace{2mm}
\noindent
\textbf{Training details} \hspace{1mm}
We use ADAM optimizer~\cite{kingma2014adam} having a learning rate of $0.001$ with minibatch-size $7$. For training the local feature descriptor network, we generate dense correspondence annotations on Keypoint-5  and Pascal3D+ dataset. Whereas, the regression network is trained using Pascal3D+ and ObjectNet3D datasets.

\begin{table}[t]
\caption{\label{table:table_1} Performance comparison of baseline ablations for different design choices.}
\centering
  \begin{tabular}{lcc}
  \toprule
  Methods & $Acc_{\pi/6}$ & \textit{MedErr} \\
  \hline
  $\textit{iSPA}_{B} (baseline)$ & 77\%  & 12.52\\
  $\textit{iSPA}_{B} + (\mathcal{F}_{I_a},\mathcal{F}_{I_b}) $ & 78.5\% & 11.26 \\
  $\textit{iSPA}_{B} +(\mathcal{F}_{I_a},\mathcal{F}_{I_b}) + CNN_D$ & 79.32\% & 10.66 \\
  $\textit{iSPA}_{B} +(\mathcal{F}_{I_a},\mathcal{F}_{I_b})+ CNN_D + AuxLoss$ & \textbf{86\%} & \textbf{8.96} \\
  \midrule
  \textit{iSPA-Net} Naive Binning & 75.5 & 13.4\\
  \bottomrule
  \end{tabular}
\label{table:table_1}
\vspace{-1pt}
\end{table}

\subsection{Ablative Analysis}
\label{ex:ablations}
In this section, we experimentally validate the improvements caused by addition of the various components in our pipeline. Our baseline evaluation of architectural modifications focuses on Chair category, as it is considered as one the most challenging class with high amount of intraclass diversity.

\vspace{2mm}
\noindent
\textbf{Ablations of \textit{iSPA-Net} pipeline:} 
Our baseline model, $iSPA_B$ uses only the correspondence map $S_{I_a,I_b}$, and is trained using the Geometric Structure Aware Loss Function. First, processed features $(\mathcal{F}_{I_a},\mathcal{F}_{I_b})$ are introduced in the pipeline. Then, the disparity network $CNN_D$ is added. Finally, The Auxiliary Loss (Section~\ref{section:33}) is appended to \textit{iSPA-Net} to complete the full pipeline. As shown in Table \ref{table:table_1}, each of these enhancements leads to increased performance of \textit{iSPA-Net}. 
In the last 2 rows of Table \ref{table:table_1}, we compare the performance of \textit{iSPA-Net} with naive uniform binning of $\delta \theta$ to \textit{iSPA-Net} with $\mu$-Law quantization of $\delta \theta$ (Same as  $\textit{iSPA}_{B} +(\mathcal{F}_a,\mathcal{F}_b)+ CNN_D + AuxLoss$). It is clear from these results that $\mu$-Law quantization of $\delta \theta$ improves the performance of \textit{iSPA-Net}.

\begin{figure}[t]
\centering    
\includegraphics[width=0.8\linewidth]{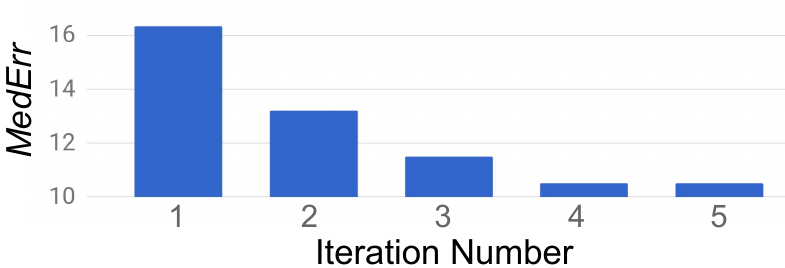}
 	\caption{\textit{MedErr} of \textit{iSPA-Net} with respect to iterative prediction limit $n$.  
 	} 
 	\label{fig:fig_iter}    
\end{figure}

\vspace{2mm}
\noindent
\textbf{Ablations on number of iterations:} 
Figure \ref{fig:fig_iter} show the improvement in the viewpoint estimation due to refinement of prediction in consecutive iterations. As is evident from the Figure, \textit{iSPA-Net}'s performance improves iteratively, supporting the notion of iterative refinement of pose for Object viewpoint estimation.

For all the ablations, the network is trained on the train-subset of ObjectNet3D and Pascal-3D+ dataset. We report our ablation statistics on the test-subset of Pascal3D+ for the chair category.

\subsection{Viewpoint Estimation}
In this section, we evaluate \textit{iSPA-Net} against other \textit{state-of-the-art} networks for the task of viewpoint estimation.

\vspace{2mm}
\noindent
\textbf{Evaluation on ObjectNet3D:} \hspace{1mm}
ObjectNet3D dataset consists of 100 diverse categories, 90,127 images with 201,888 objects. 
Due to lack of keypoint annotation, we evaluate \textit{iSPA-Net} on 4 categories from this dataset, namely, Chair, Bed, Sofa and Dining-table. To evaluate the viewpoint estimation of \textit{iSPA-Net}, we report performance in terms of $Acc_\theta$ and $MedErr$ using the ground truth bounding boxes provided with the dataset.

Due to Lack of prior work on this dataset, we additionally trained RenderForCNN~\cite{su2015render} on this dataset using the code and data provided by the authors Su~\etal. The results are presented in Table~\ref{tab:objnet_main}. RenderForCNN is observed to perform poorly on this dataset. This is due to the fact that the synthetic data provided by the authors is overfit to the distribution of Pascal3D+ dataset. The poor performance of RenderForCNN not only highlights its lack of generalizability, but also demonstrates the susceptibility of models trained on synthetic data to falter on real data even on slight mismatch of image distributions. Stricter metrics such as ($Acc_{\pi/8}$ and $Acc_{\pi/12}$) further emphasize the superiority of our method. 

\begin{table}[b] 

\vspace{2mm}

\caption{Evaluation on viewpoint estimation on ObjectNet3D dataset. Note that \textit{iSPA-Net} is trained with no synthetic data, where as Su~\etal  is trained with 500,000 synthetic images (for all 4 classes).}
\centering
\begin{tabular}{ccccccc}
  \toprule
Method& Metric&Chair  & Sofa & Table  & Bed & Avg. \\ 
\hline
 \multirow{2}{*}{$MedErr$}& Su~\etal~\cite{su2015render} &   9.70&  8.45 & \textbf{4.50} &  7.21 &  7.46 \\
& $\textit{iSPA-Net}$&  \textbf{9.15} &  \textbf{6.08} & 4.70 &  \textbf{7.11} &  \textbf{6.76} \\ 
\hline
 \multirow{2}{*}{$Acc_{\pi/6}$}& Su~\etal~\cite{su2015render} &  0.75 & 0.90 & 0.77 & 0.77 & 0.80 \\
& $\textit{iSPA-Net}$& \textbf{0.82} &\textbf{0.92} &\textbf{0.95 }& \textbf{0.83} & \textbf{0.88} \\
\hline
 \multirow{2}{*}{$Acc_{\pi/8}$}& Su~\etal~\cite{su2015render}& 0.71 & 0.89 &0.72 & 0.75 & 0.76 \\
&$\textit{iSPA-Net}$& \textbf{0.79} & \textbf{0.91} & \textbf{0.93} & \textbf{0.80} & \textbf{0.85} \\
\hline
 \multirow{2}{*}{$Acc_{\pi/12}$}& Su~\etal~\cite{su2015render}& 0.64 & 0.80 &0.68  &  0.72 & 0.71\\
& $\textit{iSPA-Net}$ & \textbf{0.67} & \textbf{0.86} & \textbf{0.89}  &  \textbf{0.74} & \textbf{0.79} \\
\bottomrule
\end{tabular}
\label{tab:objnet_main}
\end{table}

\begin{table*}[t]
\caption{Performance for object viewpoint estimation on PASCAL 3D+~\cite{xiang2014beyond} using ground truth bounding boxes. Note that \textit{MedErr} is measured in degree. }
\centering
\begin{tabular}{lccccccccc}
\toprule
\multirow{2}{*}{Category} \qquad &&
 \multicolumn{2}{c}{Su~\etal~\cite{su2015render}} \qquad &&\multicolumn{2}{c}{Grabner~\etal~\cite{Grabner18}} \qquad && \multicolumn{2}{c}{Ours $\textit{iSPA-Net}$}   \\
&& $\quad Acc_{\pi/6}$ & $ MedErr \quad$ && $\quad Acc_{\pi/6}$ & $MedErr \quad$ && $\quad Acc_{\pi/6}$ & $MedErr \quad$ \\ 
\hline
Chair &&   0.86 & 9.7  && 0.80  & 13.7  && \textbf{0.86} & \textbf{8.96}  \\
Sofa  &&  \textbf{0.90}  & 9.5  && 0.87  & 13.5  && 0.88  & \textbf{9.37}   \\
Table && 0.73  & 10.8  && 0.71  & 11.8  && \textbf{0.83}  & \textbf{6.28} \\ 
\hline
Average &&  0.83  & 10.0  && 0.79  & 13.0  && \textbf{0.86}  & \textbf{8.20}  \\ 
\bottomrule
\end{tabular}
\label{tab:pascal_main}
\end{table*}


\vspace{2mm}
\noindent
\textbf{Evaluation on Pascal3D+:} \hspace{1mm}
Pascal 3D+~\cite{xiang2014beyond} dataset contains images from Pascal~\cite{everingham2015pascal} and ImageNet~\cite{russakovsky2015imagenet} labeled with both detection and continuous pose annotations for 12 rigid object categories. Due to lack of keypoint annotation information, we show results for 3 classes, namely Chair, Sofa, and Dining-table Similar to 3D-INN\cite{wu2016single}. 

We observe in Table~\ref{tab:pascal_main} that \textit{iSPA-Net}, even with significantly less data than RenderForCNN, is able to surpass current \textit{state-of-the-art} methods. 



\begin{table}[t] 
\caption{Performance for object viewpoint estimation on PASCAL 3D+~\cite{xiang2014beyond} using ground truth bounding boxes, for stricter metrics. Note that we use $\sim$95\% less synthetic data than RenderForCNN.}
\centering
\begin{tabular}{cccccc}
  \toprule
Method& Metric&Chair  & Sofa & Table  & Avg. \\ 
\hline
 \multirow{2}{*}{$Acc_{\pi/8}$}& Su~\etal~\cite{su2015render} &   0.59&  0.76 & 0.68 &    0.68 \\
& $\textit{iSPA-Net}$&  \textbf{0.84} &  \textbf{0.80} & \textbf{0.83} &    \textbf{0.82} \\ 
\hline
 \multirow{2}{*}{$Acc_{\pi/12}$}& Su~\etal~\cite{su2015render} &  0.42 & 0.69 & 0.60  & 0.57 \\
& $\textit{iSPA-Net}$&  \textbf{0.76} &  \textbf{0.75} & \textbf{0.78} &    \textbf{0.76} \\ 
\bottomrule
\end{tabular}
\vspace{5pt}
\label{table:table_4}
\end{table}

\subsection{Favorable Attributes of \textit{iSPA-Net}}

\begin{figure}[b]

\vspace{2mm}
\centering    
\includegraphics[width=0.95\linewidth]{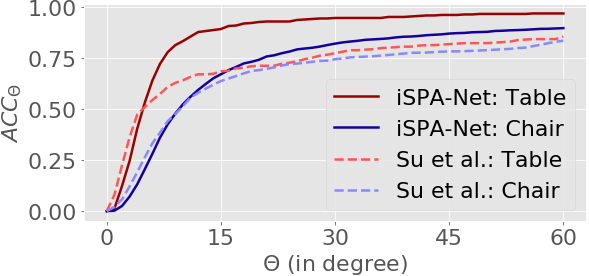}
 	\caption{Comparison of \textit{iSPA-Net} to Su~\etal for varied values of $\mathbf{\theta}$ in $\mathbf{Acc_\theta}$ metric on ObjectNet3D dataset.
    \label{fig:fig_roc}  
	}   
\end{figure}

\vspace{2mm}
\noindent
\textbf{Fine-Grained Pose Estimation}: In Table~\ref{table:table_4}, we compare \textit{iSPA-Net} to RenderForCNN on stricter metrics, $Acc_{\pi/8}$ and $Acc_{\pi/12}$. As shown in the table, \textit{iSPA-Net} is clearly superior to RenderForCNN for fine-grained pose estimation. Further, in Figure~\ref{fig:fig_roc}, we show a plot of $Acc_\theta$ vs $\theta$ on ObjectNet3D dataset. Compared to the previous \textit{state-of-the-art} models, we are able to substantially improve the performance with harsher $\theta$ bounds, indicating that our model is highly precise on estimating the pose of object in many images. Figure~\ref{fig:fig_roc}, shows $Acc_\theta$ vs. $\theta$  for two different categories in ObjectNet3D test-set. 


\begin{table}[t] 
\caption{Performance for object viewpoint estimation on PASCAL 3D+~\cite{xiang2014beyond} for single model training regime, which highlights the generalizability of \textit{iSPA-Net}'s learned representation.}
\centering
\begin{tabular}{cccccc}
  \toprule
Method& Metric&Chair  & Sofa & Table  & Avg. \\ 
\hline
 \multirow{2}{*}{$MedErr$}& Grabner~\etal~\cite{Grabner18}  &   15.90&  11.60 & 16.20 &    14.57 \\
& $\textit{iSPA-Net}_S$&  \textbf{13.56} &  \textbf{8.98} & \textbf{11.84} &    \textbf{11.45} \\ 
\hline
 \multirow{2}{*}{$Acc_{\pi/6}$}& Grabner~\etal~\cite{Grabner18} &  0.72 & 0.80 & 0.67  & 0.73 \\
& $\textit{iSPA-Net}_S$&  \textbf{0.74} &  \textbf{0.80} & \textbf{0.83} &    \textbf{0.79} \\ 
\bottomrule
\end{tabular}
\vspace{5pt}
\label{tab:general}
\end{table}

\vspace{2mm}
\noindent
\textbf{High Generalizability}: For pose estimation, the memory-efficiency of a given approach is a crucial detail for its deployment. \textit{iSPA-Net} achieves memory-efficiency by being highly generalizable across object categories. We train a single network for all the above considered objects categories, and compare it to the single network performance of Grabner~\etal~\cite{Grabner18}. In Table~\ref{tab:general}, we show that \textit{iSPA-Net} clearly outperforms Grabner~\etal. Note that the single network model of Grabner~\etal  is trained on 12 classes. However, due to the significantly better performance of our approach, we assert that our approach is equally, if not more, generalizable.

\begin{figure}[b]

\vspace{2mm}
\centering    
\includegraphics[width=1.0\linewidth]{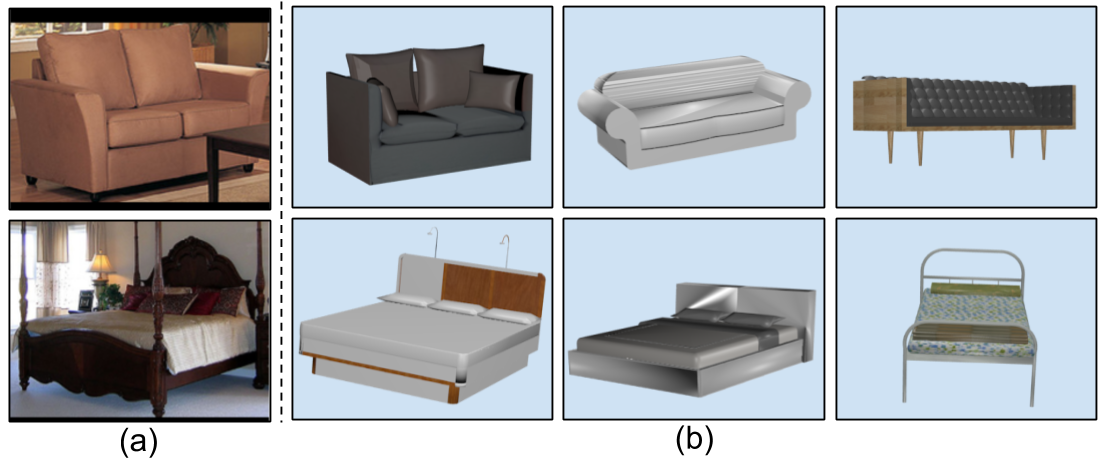}
 	\caption{Illustration of active view point localization; (a) given reference image, (b) localized view-point on various 3D objects.
 	} 
 	\label{fig:fig_4}    
\end{figure}

\section{Applications of \textit{\lowercase{i}SPA-N\lowercase{et}}}

\begin{figure*}[!t]
\centering    
\includegraphics[width=0.95\linewidth]{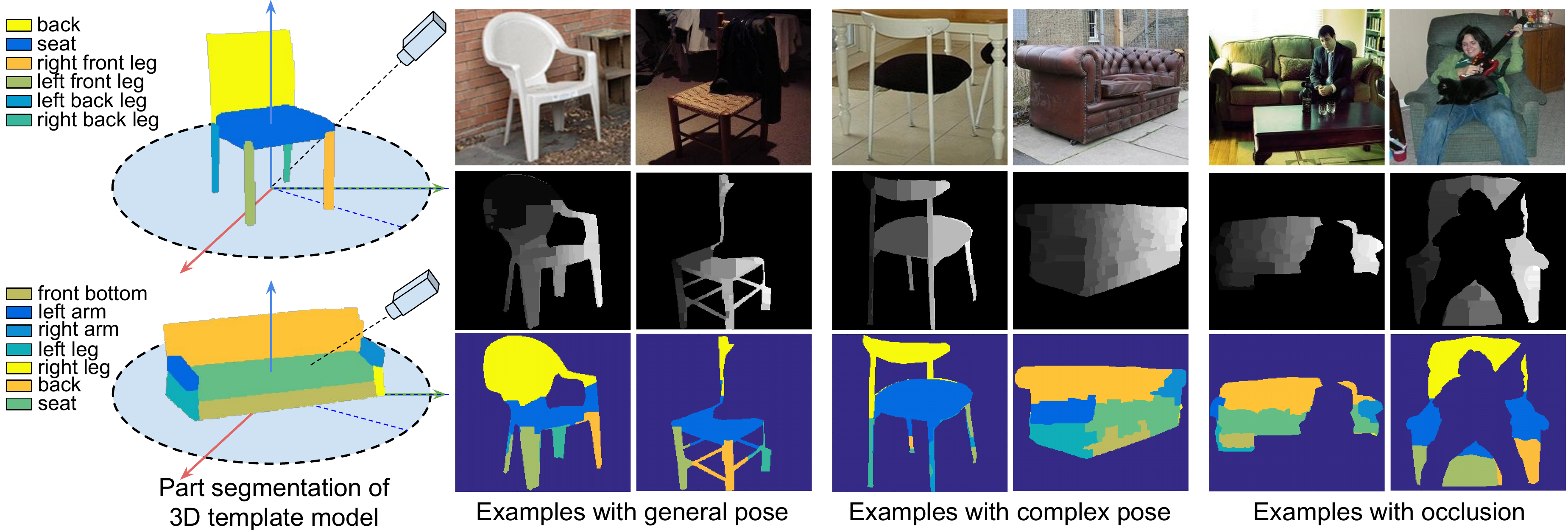}
 	\caption{Qualitative results of unsupervised semantic part segmentation. First, \textit{iSPA-Net} performs pose-alignment, which is then followed by transfer of part labels from the template model to the given real image. Note that only a single template model has to be annotated per category.
 	} 
 	\label{fig:fig_5}    
\end{figure*}

\subsection{Object viewpoint Localization} \label{section:41}
In this section we present a novel application of \textit{iSPA-Net}. As presented earlier, object viewpoint localization is the task of estimating the location, from where pose of object in 3D world can match the pose of object in a given reference image. Solutions for this task can be used for multiple industrial applications such as automated object cataloging, massive manufacturing survey etc.  \textit{iSPA-Net} is designed to estimate fine-grained pose of objects without relying on the presence of a similar CAD model for alignment. Our model can be fixed on a drone, which receives a reference image $I_b$ and has an input feed (e.g. a camera) to obtain real world image $I_a$. Now, instead of rendering $I_b'$ at a new viewpoint for pose alignment, the drone can maneuver to a different location giving rise to a updated camera-feed $I_a'$, so as to align with the pose in $I_b$ in consecutive iterations. 

Due to lack of experimental setup, we qualitatively evaluate this task on synthetic dataset. Using \textit{iSPA-Net}, we align different 3D objects to a single given reference image $I_b$. Figure \ref{fig:fig_4} shows our qualitative results for this task, where images in (b) show localized viewpoints for various 3D objects based on the reference image given in (a).

\subsection{Semantic Part Segmentation}  \label{section:42}
As an additional application of the proposed pose alignment framework, we perform  semantic part segmentation transfer from the chosen template model to multiple real images containing instances of the same object class ~\cite{huang2015single}. We manually annotate only the 3D template models (one for each object class) used by the pose alignment network for each object class as shown in Figure \ref {fig:fig_5}. 
To transfer the part segmentation from the annotated template model, we first perform pose alignment using the proposed \textit{iSPA-Net} framework. From the pose aligned render of the 3D-template projection (with labels),  feature-correlation from the intermediate correspondence network output is utilized for semantic label transfer. This is done by assigning a label to each pixel in the real image based on the label of the spatial features in the 3D-template projection which are highly correlated to features of that pixel. Then, a silhouette map for the given natural object images is obtained using state-of-the-arts object segmentation model~\cite{chen2017rethinking}. A hierarchical image segmentation algorithm as proposed by Arbelaez \etal~\cite{arbelaez2011contour}  is then used to acquire super-pixel regions in the image. The over-segmented regions are obtained only for the region masked by the silhouette map obtained from the segmentation model as shown the second row of each examples in Figure~\ref{fig:fig_5}. For each over-segmented region we assign median value of the comprising pixel-labels obtained from the correlation based label-transfer step. The resultant part-segmentation map is shown in the last row of Figure \ref{fig:fig_5}. 

It is evident from the qualitative results that the pose-alignment network can be used effectively to obtain a coarse level part segmentation even in presence of diverse view and occlusion scenarios (see Figure \ref{fig:fig_5}). Such pose-alignment based approach also opens up possibilities to improve the available part-segmentation models by utilizing fine-grained pose information in a much more explicit manner. Moreover, use of pose-alignment to obtain part-segmentation can be used to assist annotators with an initial coarse label map. The procedure involves manual segmentation of a single template model per class, which also addresses the scalability issue of part-segmentation algorithms for novel object categories.


\section{Conclusions}
In this paper, we present a novel iterative object viewpoint estimation network, \textit{iSPA-net},  for fine-grained Pose estimation, drawing inspiration from human perception and classical computer vision pipeline. Along with demonstrating \textit{state-of-the-art} performance in various public datasets, we also show that such a pipeline can have wide industrial applications. This work presents a multitude of new challenges as well, such as formulating an unsupervised approach for annotation-free training regime, estimating pose of diverse outdoor-and-indoor objects etc. Along with facing the aforementioned challenges, our future work will focus on extending the proposed framework to perform 6D object pose tracking.

\vspace{2mm}
\noindent
\textbf{Acknowledgements}\hspace{2mm} This work was supported by a CSIR Fellowship (Jogendra), and a project grant from Robert Bosch Centre for Cyber-Physical Systems, IISc.
\pagebreak

\title{Supplementary: iSPA-Net: Iterative Semantic Pose Alignment Network
\vspace{15mm}
}

\makesuptitle

In this supplementary we outline the various secondary details which provide interesting insight into this work, while also elaborating on various intricacies of our approach.

\section{Data Generation}
\label{sec:data}
\subsection{Overview}
We select a single 3D template model $M_c$ for each class of object $c \in C$ from ShapeNet dataset \cite{shapenet2015}. 
Using a modified version of the rendering pipeline presented by  \cite{su2015render}, we render the selected template model $M_c$ at various viewpoints to create samples of image $I_2(\theta_2, \phi_2, \psi_2 )$. 
Note that, our reliance on synthetic data is minimal. We use only 8K renders of  a single 3D template model. This is in sharp contrast to works such as \cite{su2015render,li2017deep}, which use millions of synthetic images in their training pipeline.

\begin{figure}[h]
\centering    
	\includegraphics[width=0.95\linewidth]{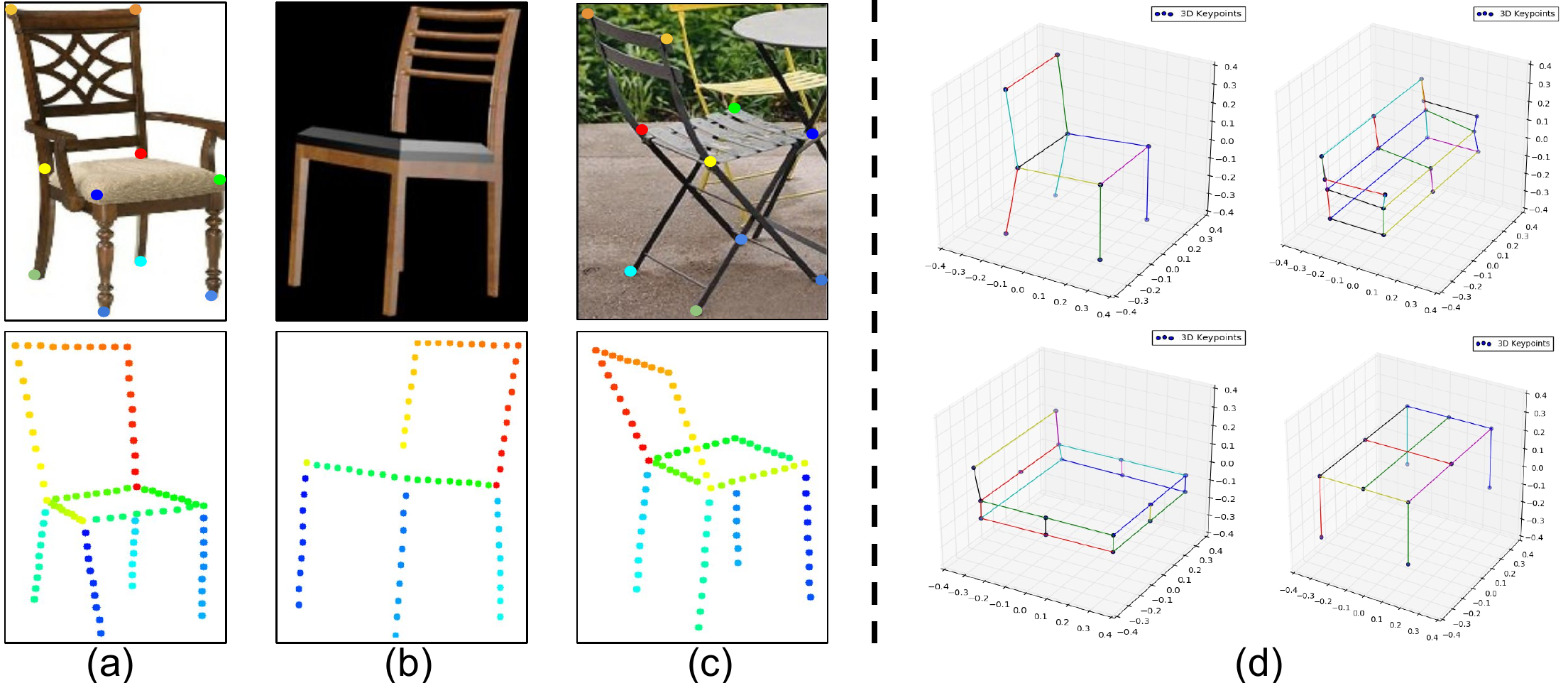}
	\caption{Data Generation. 
	}
	\label{fig:dat_gen}    
\end{figure}

To train our pose-invariant local descriptors, $\mathcal{L}$, we use the contrastive corresondence loss function, introduced in\cite{choy2016universal}, which requires dense correspondence annotations between image pair $(I_1,I_2)$. For generating such correspondence information, we use automated processing of annotations provided in the Keypoint-5 dataset, released in 3D-INN \cite{wu2016single}. We use the 2D skeletal representation, which is based on annotation of sparse 2D skeletal keypoints of real images. As shown in Figure \ref{fig:dat_gen}, the sparse 2D keypoints are annotated on real images at important joint locations such as leg ends, seat joint etc. For each image sample $I_1$, we join these sparse keypoints in a fixed order, to obtain the corresponding 2D skeletal frame $S_a$ (second row of Figure \ref{fig:fig_1}. a, c). 
To generate a similar skeletal frame for our rendered 3D object templates in $I_2$, we manually annotate sparse 3D keypoints for our template objects, as shown in Figure \ref{fig:dat_gen} d. Using the projection of these 3D keypoints, based on the viewpoint parameters, we generate 2D keypoints for any rendered image $I_2$ and also the corresponding  2D skeletal frame $S_2$ in a similar fashion. 
Now, by pairing points along this generated skeletal frames $(S_1,S_2)$ of any image pair $(I_1,I_2)$, 2D keypoints on $I_1$ can be matched to the corresponding keypoints on image $I_2$. Following this, we generate dense correspondence set for any image pair $(I_1,I_2)$ to improve performance of local descriptor $\mathcal{L}$.

Figure \ref{fig:dat_gen} shows some qualitative examples of generated annotations for our training. We employ various methods to prune and reform our generated annotations, such as depth based pruning (for $I_1$), and other methods, explained further in the next section. Although, sometimes the generated annotations are not accurate (Figure \ref{fig:dat_gen}.c), the correspondence model is able to learn improved view invariant local descriptors $\mathcal{L}$, due to the presence on ample amount of correct noise-free annotations.

\begin{figure}[!t]
\centering    
	\includegraphics[width=0.95\linewidth]{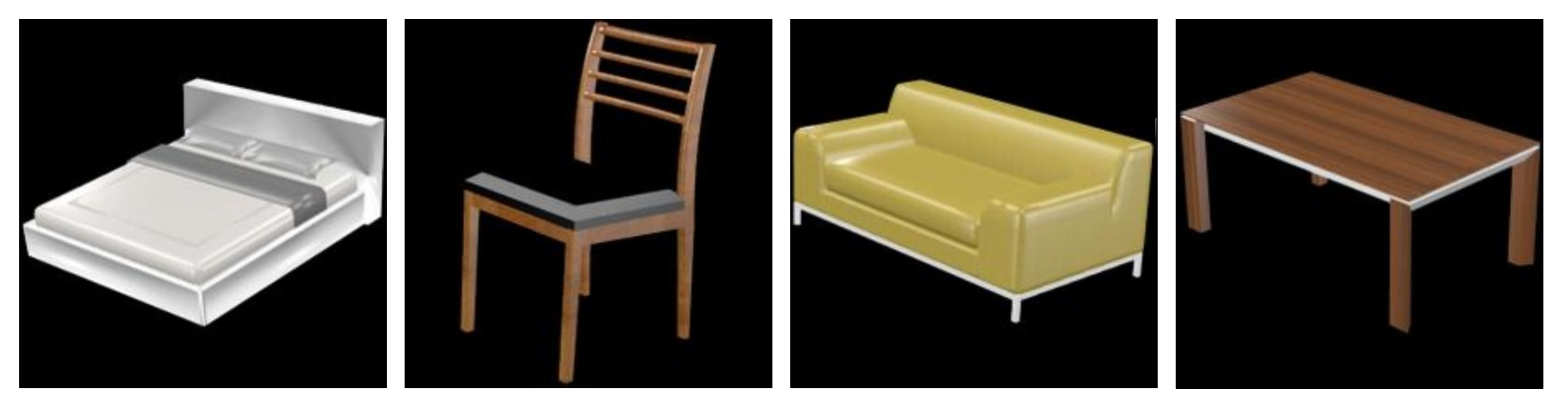}
	\caption{The template 3D model we use for each class. 
	}
	\label{fig:template}    
\end{figure}

Finally, Figure~\ref{fig:template} shows the single 3D template model used for each object category.

\subsection{Keypoint Pruning Mechanism}
\label{sub:sub}
For Keypoint pruning, we use three main approaches:

\begin{figure}[b]
\centering    
	\includegraphics[width=0.95\linewidth]{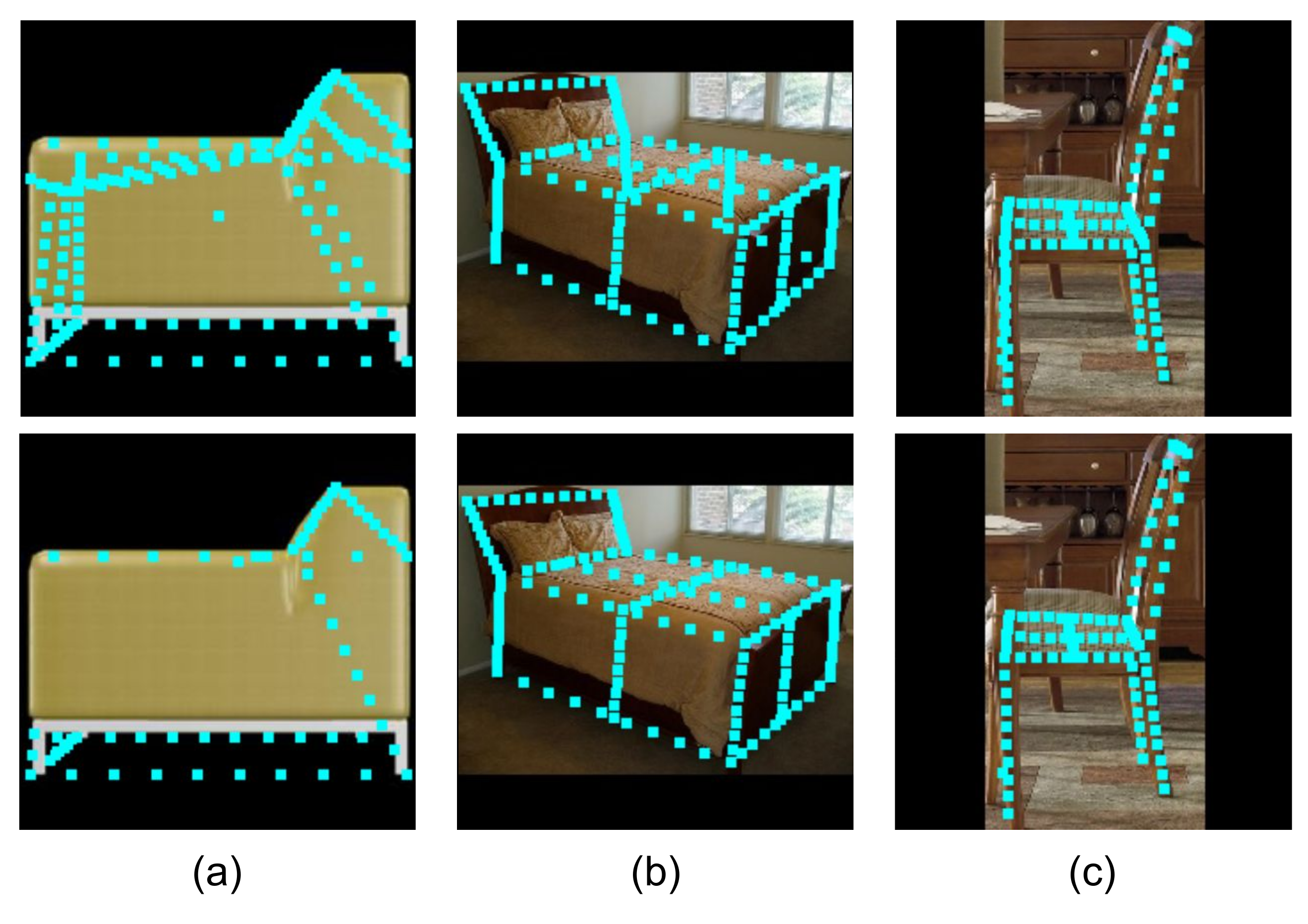}
	\caption{The utility of the three pruning mechanism presented in section \ref{sub:sub}. 
	}
	\label{fig:prune}    
\end{figure}

\begin{itemize}
    \item \textbf{Visibility Based Pruning in image $I_2$}: As the image $I_2$ is rendered using a template 3D model, a visibility map of the entire object can also be formed easily. Using this visibility map, we prune our points which are not visible from the rendered viewpoint. In figure \ref{fig:prune} (a), an example is presented.
    \item \textbf{Seat Presence in Image $I_1$}: As we know visibility information of all parts of the real image is not available. Hence, we instead use some approximations. We assume that all images of the object are from positive elevation angles. If this assumption holds true, all the leg skeletal keypoints which occur inside the the 2D region covered by the seat are not visible and hence can be pruned out. In figure \ref{fig:prune} (b), examples of this pruning mechanism is presented.
    \item \textbf{Self-Occlusion of Legs in Image $I_2$} Self occlusion of object legs can be a very frequent occurrence, and almost in all angles, some legs of an object may occlude other legs. We further prune out keypoints on occluded leg, by applying a heuristic approach. First, We approximate the pose quadrant of the object by joining a 2D vector from the back of the seat to the front. Now, based on the pose of the object, which leg can occlude the other is known. This information is then used to prune out self-occluded leg keytpoints. In figure \ref{fig:prune} (c), an example is presented.
\end{itemize}

\begin{figure}[t]
\centering    
	\includegraphics[width=0.95\linewidth]{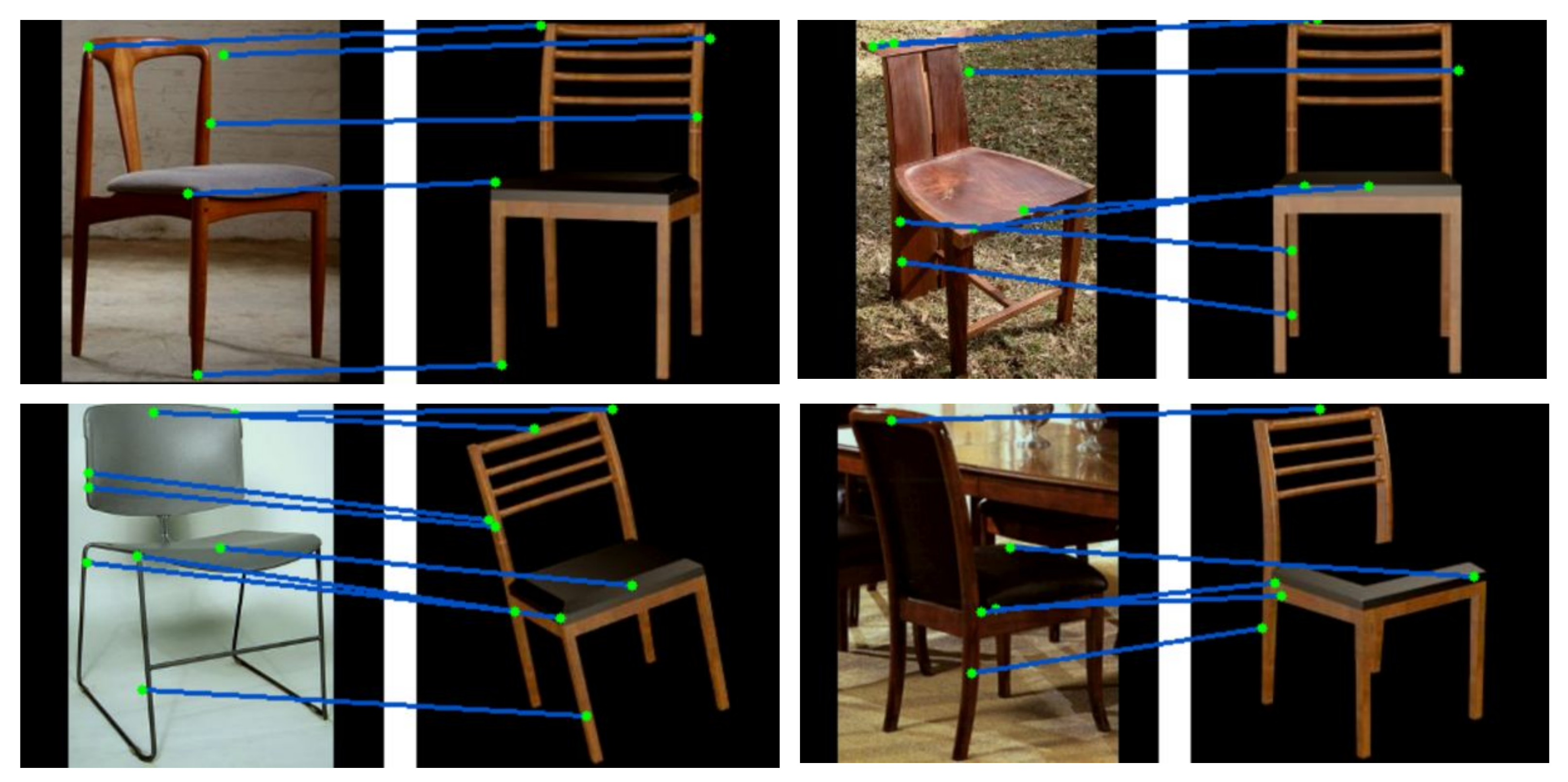}
	\caption{Keypoint Estimation results.  
	}
	\label{fig:keypoint}    
\end{figure}

\section{Keypoint Correspondence}
\label{sec:corresp}

For our proposed approach, the optimality of the learnt local descriptors for giving correspondence map is crucial. In this section, we show some qualitative results to demonstrate keypoint estimation ability of our pose-invariant local descriptors. For each keypoint in the synthetic render, we find the closest matching location in a given real image. In figure \ref{fig:keypoint}, we show some qualitative results of keypoint matching between real images and multiple renders of our template 3D model.As we can see, the learnt local descriptors are indeed pose-invariant as they are correctly corresponding to right locations even after considerable change in pose ( for example, the bottom right pair).


\begin{figure}[b]
\centering    
	\includegraphics[width=0.95\linewidth]{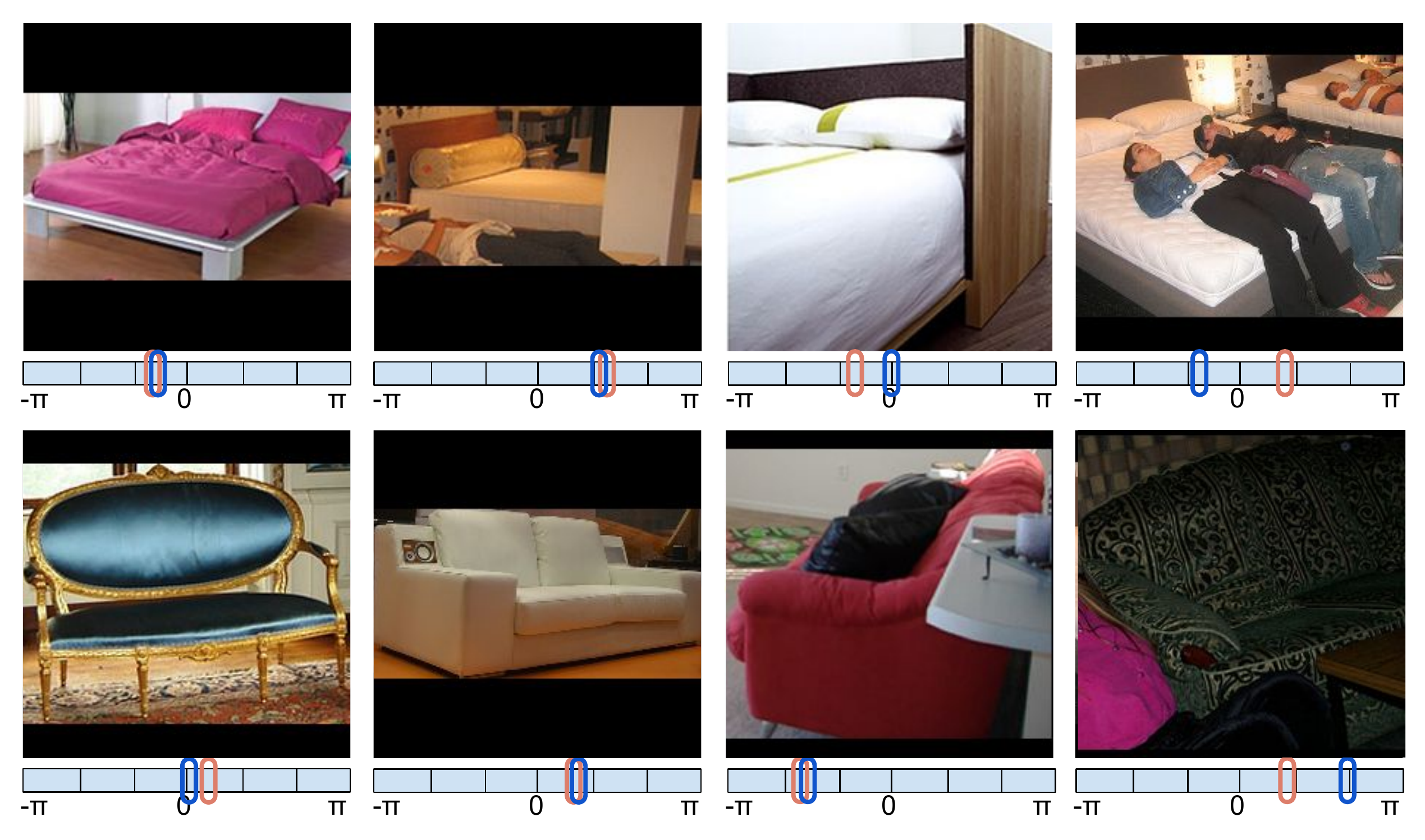}
	\caption{Pose Estimation results. The bar below each image represent the azimuth angle values. The green ellipse represents the Ground Truth pose, and the blue ellipse represents the predicted pose. 
	}
	\label{fig:pose_estimation}    
\end{figure}
\section{Qualitative Samples for Pose estimation}
\label{sec:res}
In this section, we show some of the results achieved by our network. In Figure~\ref{fig:pose_estimation}, we show examples of images from Pascal 3D+ easy test-dataset, along with predicted and annotated azimuth pose angle. The images are arranged in ascending order of angular error in azimuth estimation. As we can see, many times, high error in pose estimation occurs due to extremely poor image quality, due to factors such as lack of illumination, clutter etc.

\bibliographystyle{ACM-Reference-Format} 
\bibliography{sample-bibliography}

\end{document}